\documentclass[letterpaper, 10 pt, conference]{ieeeconf}  

\IEEEoverridecommandlockouts

\usepackage{cite}
\usepackage{amsmath,amssymb,amsfonts}
\usepackage{algorithmic}
\usepackage{graphicx}
\usepackage{textcomp}

\usepackage{color, colortbl}
\definecolor{TableColor1}{rgb}{0.93,0.90,0.95}

\definecolor{TableColor2}{rgb}{0.34,0.023,0.55}
\definecolor{TableColor3}{rgb}{0.67,0.50,0.77}
\definecolor{TableColor4}{rgb}{0.54,0.0,0.88}
\definecolor{TableColor5}{rgb}{0.43,0.17,0.62}
\definecolor{TableColor6}{rgb}{0.48,0.35,0.65}
\def\BibTeX{{\rm B\kern-.05em{\sc i\kern-.025em b}\kern-.08em
    T\kern-.1667em\lower.7ex\hbox{E}\kern-.125emX}}

\title{\LARGE \bf {Upper-limb Geometric MyoPassivity Map for Physical \\Human-Robot Interaction}}

\author{Xingyuan Zhou$^{*}$,~\IEEEmembership{Student Member,~IEEE,} Peter Paik$^{*}$,~\IEEEmembership{Student Member,~IEEE,}\\ S. Farokh Atashzar$^{\dagger}$,~\IEEEmembership{Senior Member,~IEEE}\vspace{-0.6cm}
\thanks{Xingyuan Zhou and Peter Paik are with the Department of Electrical and Computer Engineering, New York University (NYU), New York, NY, 11201 USA. Atashzar is with the Department of  Electrical and Computer Engineering, Mechanical and Aerospace Engineering, Biomedical Engineering, NYU. Atashzar is also with NYU WIRELESS Center and NYU CUSP. This material is based upon work supported by the US National Science Foundation under grants no \#2121391 and \#2208189. The work is also partially supported by NYUAD Center for Artificial Intelligence and Robotics (CAIR) award \# CG010.}\\
\thanks{$^{*}$ Xingyuan Zhou and Peter Paik contributed equally to this work and share
the first authorship.}\\
\thanks{$^{\dagger}$ Corresponding author: Atashzar ({\tt\footnotesize f.atashzar@nyu.edu}).}
}

\begin{document}

\maketitle
\thispagestyle{empty}
\pagestyle{empty}
\bstctlcite{IEEEexample:BSTcontrol}

\begin{abstract}
The intrinsic biomechanical characteristic of the human upper limb plays a central role in absorbing the interactive energy during physical human-robot interaction (pHRI). We have recently shown that based on the concept of ``Excess of Passivity (EoP)," from nonlinear control theory, it is possible to decode such energetic behavior for both upper and lower limbs \cite{a1,a2}. The extracted knowledge can be used in the design of controllers (such as \cite{a2,a3,a4,a5}) for optimizing the transparency and fidelity of force fields in human-robot interaction and in haptic systems. In this paper, for the first time, we investigate the frequency behavior of the passivity map for the upper limb when the muscle co-activation was controlled in real-time through visual electromyographic feedback. Five healthy subjects (age: 27$\pm$5) were included in this study. The energetic behavior was evaluated at two stimulation frequencies at eight interaction directions over two controlled muscle co-activation levels. Electromyography (EMG) was captured using the Delsys Wireless Trigno system. Results showed a correlation between EMG and EoP, which was further altered by increasing the frequency. The proposed energetic behavior is named the Geometric MyoPassivity (GMP) map. The findings indicate that the GMP map has the potential to be used in real-time to quantify the absorbable energy, thus passivity margin of stability for upper limb interaction during pHRI.

\end{abstract}


\section{Introduction}
Physical human-robot interaction (pHRI) is a central factor in several fields of robotics, including haptics, telerobotics, and rehabilitation robotics. In the aforementioned research areas, a critical goal is to optimize the interaction between humans and robots to (a) maximize safety and (b) manage the energy exchange between human biomechanics and robot mechanics. In the field of haptics and telerobotics, this concept closely relates to the transparency and stability of the system. The stability of pHRI is often challenged by sources of nonpassive energy such as delay, jitter, and packet loss in telerobotic systems \cite{i1,i2,i3}, besides sensor noise or actuator fault in generic pHRI systems \cite{i4}. The situation is even more challenging in some applications that require high-force interaction and energy augmentation. Examples include (tele)robotic rehabilitation and exoskeletons for patients with neural damages (such as those caused by a stroke) where the robot needs to inject assistive energy (which is mathematically nonpassive) into the high-gain closed-loop system to deliver the required assistance/therapy \cite{a5,tdpa0,stiff1,stiff2,stiff3}.

As a result, passivity control theory has often been used to detect and observe the deviation from a passivity condition and accordingly tune control parameters (such as injected damping) to impose passivity, thus stability towards safety. However, excessive energy dissipation in controllers would sacrifice the system’s performance (such as motion tracking in pHRI or force transparency in haptics-enabled telerobotics). As a result, any algorithmic solution that can support higher stability and less conservative modulation of energy simultaneously can significantly contribute to the applicability of such technologies.

One of the state-of-the-art stabilizers in pHRI systems is the Time-Domain Passivity Approach (TDPA), which guarantees the stability of the system by injecting damping to keep the observed energy levels of the closed-loop system positive and thus impose passivity and ensuring stability \cite{tdpa1,tdpa2,tdpa3,tdpa4,tdpa5}. TDPA has been widely used for haptics and networked robotics systems. In the context of networked robotics (such as telerobotics and telehaptics), TDPA ensures that the system remains stable in the face of variable delays and unknown dynamics at the cost of the ideal performance. However, one parameter which has often been ignored in the design of such controllers for pHRI is the energy absorption capacity of human biomechanics during the interaction. We have recently shown that even a conservative minimum lower bound on such a feature can significantly enhance the performance of the system when compared to state-of-the-art controllers (please see \cite{a1,a2,a3,a4,a5}).

However, quantifying such a feature is not straightforward due to: (a) the nonlinearity of human biomechanics and (b) the dependencies of such behavior on various interactional parameters. Using the concept of excess of passivity (EoP), taken from nonlinear control theory, it is possible to quantify such nonlinear energetic behavior for both upper and lower limbs \cite{a1,a2,a3,a4,a5}.

In this paper, we extend our understanding of the energetic behavior of human biomechanics (upper limb). For the first time, we evaluate the compounded effect of (a) frequency, (b) geometric direction, and (c) electromyographical muscle activation on the energy absorption capacity of human biomechanics. The outcome of the work is a frequency-aware energetic map that can be used in real-time to estimate the current EoP called Geometric MyoPassivity (GMP) Map.
The map is generated through an experiment conducted on five healthy subjects in eight different geometric directions. The results show that the energetic capacity increases with muscle activation measured by electromyography (EMG) (thus MyoPassivity). This relationship becomes stronger as the frequency of interaction decreases, as observed in all directions of the GMP map. This map can be incorporated into the design of advanced controllers to optimize the flow of energy and minimize the injection of energy dissipation using this reliance on existing biomechanical energy tanks in the system. The rest of this paper is organized as follows. In Section II, we provide the mathematical background and the experimental methodology utilized in this paper. In Section III, we provide the results and discussion about the estimated GMP map. The paper is concluded in Section IV.

\section{Preliminaries}

Taking advantage of the weak passivity theory, we can define the stability condition of the system utilizing the passivity definition\cite{def1,def2,def3,def4}.

\textbf{\textit{Definition 1 (System):}}  a system $S(t)$, with input $U(t)$, output $Y(t)$, and initial energy $E(0)$, is passive if and only if:

\begin{equation}
E_{S}(t) + E(0) \geq 0 \text{,  } \forall t > 0, 
\end{equation}\par
 
where $E_{S}(t)$ is the system's energy and is defined as:

\begin{equation}
E_{S}(t)= \int_{0}^{t}U(t)^{T}Y(t)dt.
\end{equation}\par

\textbf{\textit{Definition 2 (Interconnected System):}}
suppose the system is an interconnection of `\textit{N}' arbitrary subsystems `$S_i$'; the whole interconnection is passive if:
\begin{equation}
E_{Tot}(t) + E(0) \geq 0 \text{, } \forall t > 0,
\end{equation}\par

where the total energy of the interconnected system $E_{Tot}$ is defined as $E_{Tot}(t) = E_{1}(t) + E_{2}(t) + E_{3}(t) ... E_{N}(t) $, and $E_{i}(t)$ is the $i^{th}$ subsystem. In the context of this paper and for pHRI, the interconnected system is composed of the human biomechanics as one subsystem and the robot (generating the interactive force field) as a different subsystem.

It should be noted that, in practice and literature, the user’s limb is assumed to be passive due to its intrinsic mechanical characteristic \cite{hand1,hand2,hand3,hand4}. Therefore, the stability of the interconnected system depends on the degree of passivity that exists in human biomechanics and the degree of nonpassivity that the robotic system generates from its force field.

\textbf{\textit{Definition 3 (Output Strictly Passive System):}} a system with input $U(t)$, output $Y(t)$, and initial energy $E(0)$ is deemed output strictly passive (OSP) if it meets the following condition:

\begin{equation}
\int_{0}^{t}U(t)^TY(t)dt + E(0) \geq  \xi\int_{0}^{t}Y(t)^TY(t)dt
\end{equation}

with an EoP coefficient of $\xi \geq 0$. It has been shown that a strictly passive system is also asymptotically stable, while an OSP system is L2 stable with a finite L2 gain of $1/\xi$. It should also be mentioned that considering the definition given in (4), if $\xi < 0$, then the system is Output Non-Passive (ONP) with an Shortage of Passivity (SoP) coefficient of $\xi$. 

In the context of pHRI and using the above-given definitions from nonlinear control theory, EoP of human biomechanics describes the energetic capacity of the human limb in absorbing the interactional energy while relaxing the assumption of linearity on the biomechanics. The higher the EoP, the higher the passivity margin of the interconnected system. Based on the definitions above, for pHRI, we have shown if the EoP of the user’s upper-limb biomechanics is greater than the SoP of the force field injected by the robot into the interconnection, the overall pHRI system will remain L2 stable \cite{tdpa0}. As a result, it can be of paramount importance to estimate the EoP of the human in real-time to be used by stabilizers (such as \cite{a2,a3,a4,a5}) with the goal of closed-loop passivity observation and minimal stabilization (i.e., guaranteeing the stability while minimizing the conservatism). However, calculating the biomechanical EoP requires offline perturbation studies to decode this nonlinear feature. Thus, it is critical to learn the predictive factors of interactions which can be used in real-time to estimate the current EoP. In this paper, we study the correlation between EoP and (a) muscular co-activation, (b) frequency of interaction, and (c) geometric direction of task conduction.

\begin{figure}[htbp]
\centerline{\includegraphics[width=0.48\textwidth]{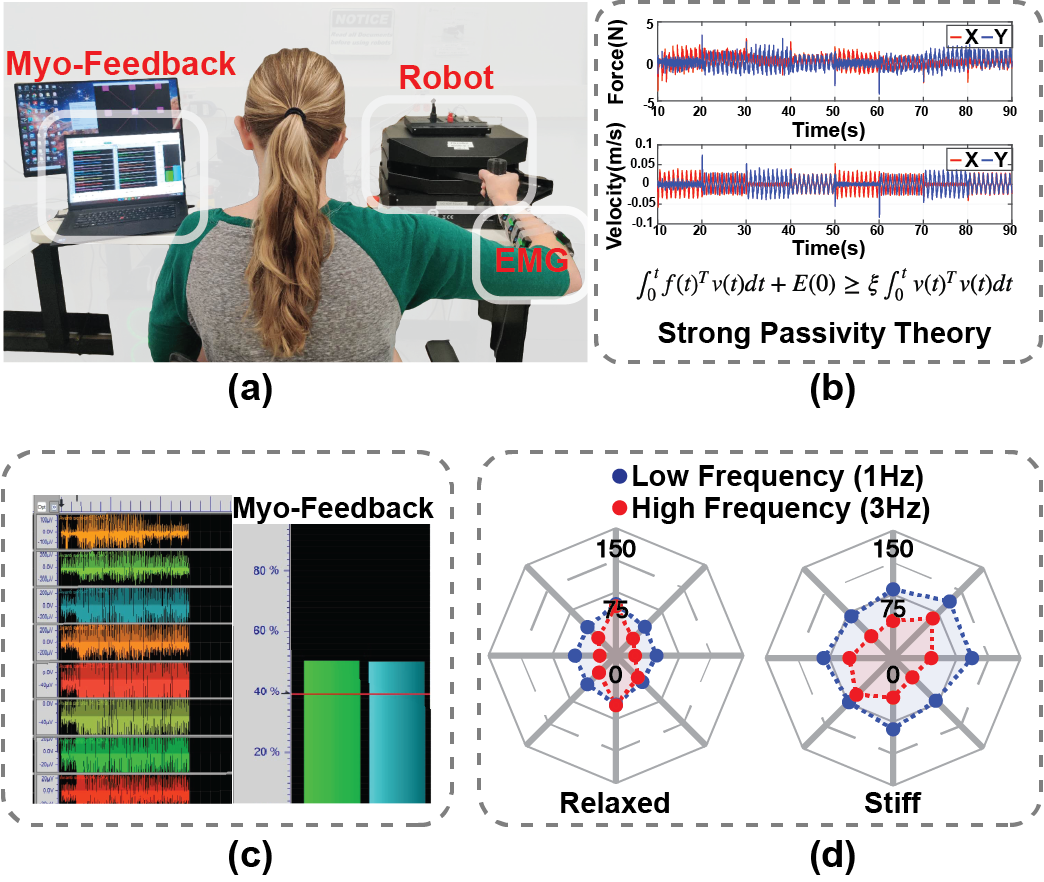}}
\caption{(a) Experimental Set-up. (b) Example of reactive force and velocity profiles. (c) Real-time EMG recording and visual myo-feedback. (d) Resulting GMP maps.}
\label{fig3}
\end{figure}

\section{Methodology}

\subsection{Experimental Set-up}
An offline identification experimental study is conducted to identify the EoP of the human biomechanics during pHRI. The experimental set-up is shown in Fig. \ref{fig3}. In the experiment, the subjects are asked to hold the robot handle and let the robot perturb their limb under different designed conditions over eight directions at two separate perturbation frequencies. During the perturbations, all required force, motion, and muscle activation data are collected to (a) calculate the biomechanical EoP and (b) investigate the correlation between interactional factors (EMG, frequency of interaction, geometric direction) and the calculated EoP. 

\begin{figure}[!h]
\centerline{\includegraphics[width=0.35\textwidth]{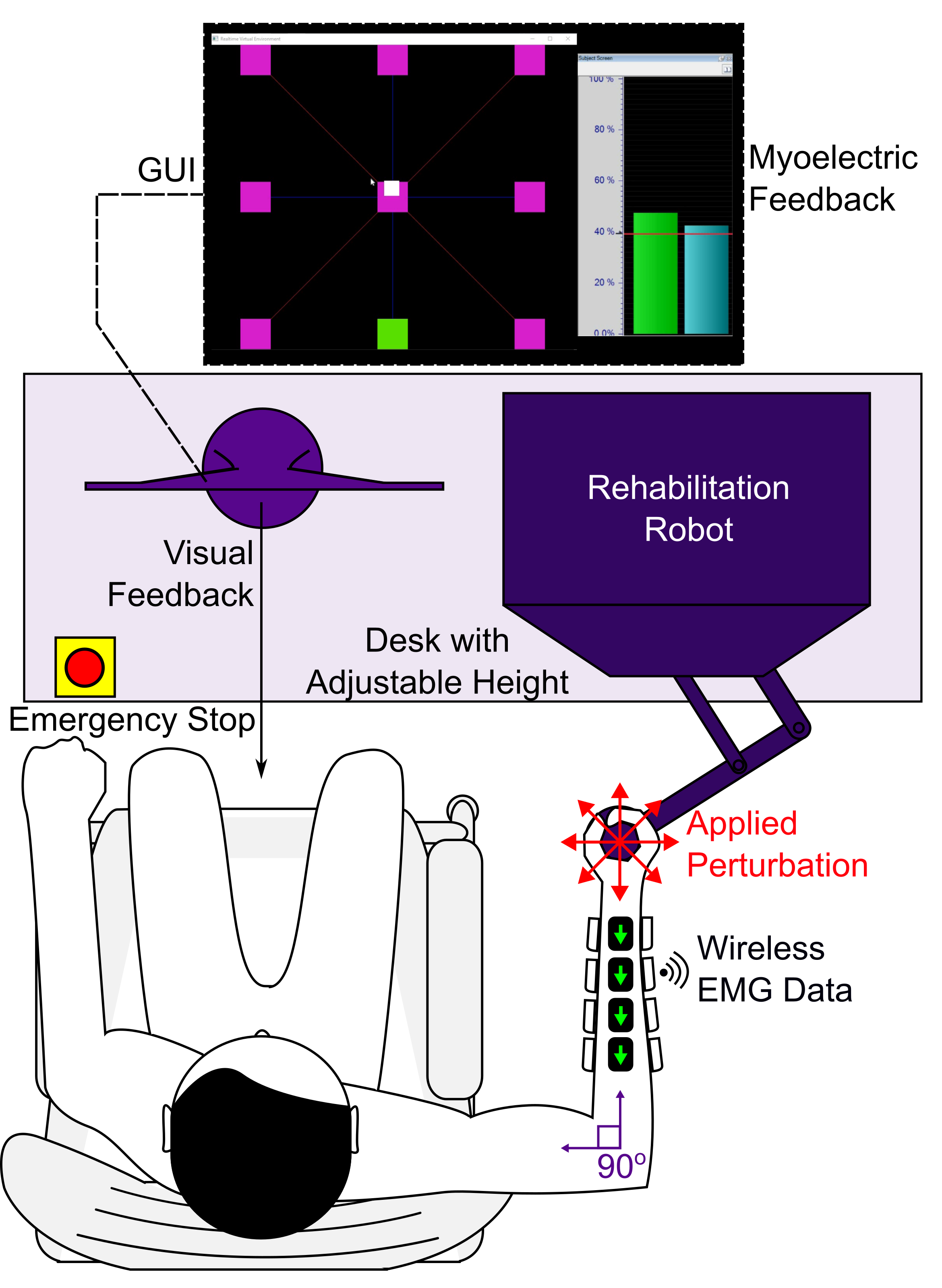}}\vspace{-0.2cm}
\caption{Experimental set-up showing the posture of the subject, rehabilitation robot, GUI, myoelectric bar feedback, and Delsys wireless system.}\vspace{-0.2cm}
\label{set-up}
\end{figure}

To accomplish this goal, a standard protocol and apparatus are designed and used for each subject (see Fig. \ref{set-up}). The experimental apparatus includes (a) a robotic system to provide the perturbation and collect the motion and force data, (b) an EMG system to collect the muscle activations, (c) visual myo-feedback to provide sensory awareness to the user during the data collection regarding the co-activation of the muscles, and (d) a height-adjustable table to control the posture of the user.

The robot used for perturbation is a Quanser 2-DOF Rehabilitation Robot (Quanser, Markham, ON, Canada), capable of perturbing the user’s biomechanics in X-Y directions and simultaneously measuring the subject’s corresponding motion and force reaction. The system used for collecting the EMG is a sixteen-channel wireless Bipolar Delsys Trigno system (Delsys, Natick, MA, USA). EMG electrodes are attached to the user’s dominant forearm and measure the muscle activity in real-time. The bipolar EMG electrodes measure the raw EMG data at 2148Hz. Four electrodes are placed along each of the muscles: extensor digitorum, extensor carpi radialis, palmaris longus, and flexor carpi ulnaris. The EMG data provides visual myo-feedback for the user to control the task as prescribed by the protocol, considering two levels of muscle co-activation (explained later). 

In this experiment, a height-adjustable desk was also used that controls the robot’s distance from the ground, thus allowing for implementing the same prescribed posture for all users while estimating the EoP. Before the experiment, the maximum voluntary contraction (MVC) of two electrodes is measured. One of the electrodes is from the palmaris longus, and the other is from the extensor digitorum. These two muscles are chosen due to their sensitivity to co-activation in the forearm, especially in pHRI. During the experiment, the subject is provided with visual feedback comprised of (a) a graphical user interface that shows the position of the end-effector besides (b) a myo-feedback in the format of two bars showing the percent maximum voluntary contraction (\%MVC).

During the experiments, the subjects are asked to position the chair so that their upper arm and forearm are perpendicular when reaching the central position, as shown on the GUI. Throughout the experiment, subjects are asked to hold the handle while minimizing any voluntary motion to the handle while letting the robot perturb the hand.

\subsection{Experimental Procedure}
The experiment is divided into two stages: (1) estimation of maximum voluntary contraction and (2) conduction of four randomly-ordered experiments (combinations of two levels of co-activation and two levels of perturbation frequency). In the first stage, the subject is asked to grasp the handle as hard as they can for three seconds, followed by a rest, and repeated a second time. After this phase, the visual myo-feedback is generated in the format of \%MVC, which is the RMS value of the EMG recordings normalized to their corresponding recorded MVC values. In phase 2, four tests are conducted in a randomized sequence: low-frequency relaxed (LR), low-frequency stiff (LS), high-frequency relaxed (HR), and high-frequency stiff (HS). For low-frequency conditions, a 1 Hz sinusoidal perturbation is applied along eight cardinal directions. And for high-frequency conditions, a 3 Hz sinusoidal perturbation is applied. The perturbation duration in each direction is 10 seconds. For relaxed conditions, the subject is asked to hold the handle with minimal muscle activation. For stiff conditions, the subject is asked to maintain a 40\% MVC following a reference provided by the visual myo-feedback.

\begin{figure*}[htbp]
\vspace{0.2cm}\centerline{\includegraphics[width=0.95\textwidth]{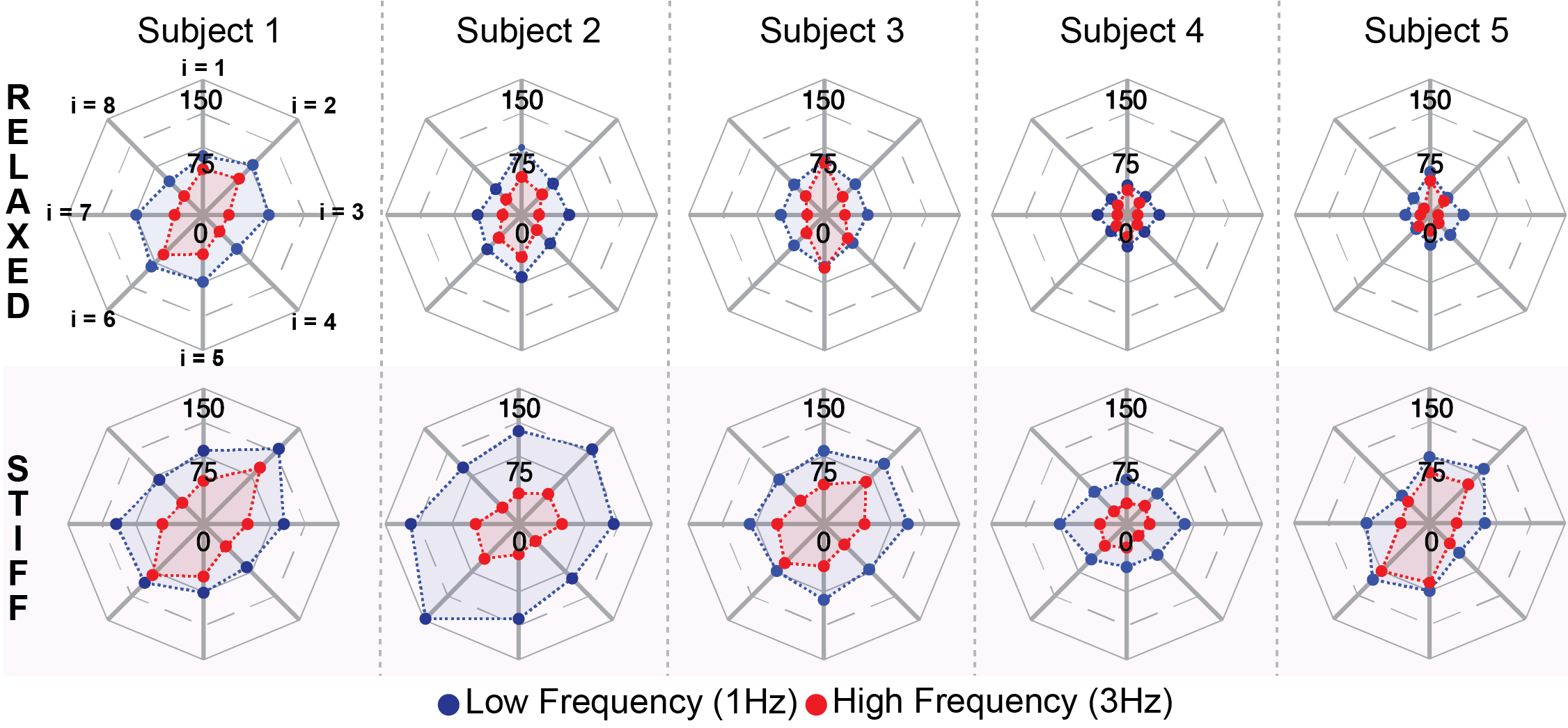}}
\caption{Resulting GMP Map for all subjects during relaxed grip (top) and stiff grip (bottom).}
\label{maps}
\end{figure*}

\begin{table}[!h] 
\normalsize
\renewcommand{\arraystretch}{1.3}
\caption{Demographic Data}
\label{Table1}
\centering
\begin{tabular}{|c|c|c|c|c|}
\hline
\rowcolor{TableColor6} 
\textcolor{white}{\textbf{Subject}} & \textcolor{white}{\textbf{Height (m)}} & \textcolor{white}{\textbf{Weight (kg)}} & \textcolor{white}{\textbf{Age}} & \textcolor{white}{\textbf{Sex}}\\
\hline

\text{1} & \text{1.87} & \text{77} & \text{32} & \text{M}\\
\hline
\rowcolor{TableColor1} 
\text{2} & \text{1.62} & \text{58} & \text{24} & \text{M}\\
\hline

\text{3} & \text{1.77} & \text{64} & \text{24} & \text{M}\\
\hline
\rowcolor{TableColor1} 
\text{4} & \text{1.70} & \text{57} & \text{24} & \text{F}\\
\hline
 
\text{5} & \text{1.80} & \text{75} & \text{23} & \text{M}\\
\hline
\end{tabular}
\end{table}

\subsection{Participants}
Five healthy subjects (four males, one female, 27 ± 5 years) participated in the study. The institutional review board of New York University approved the study, and subjects provided their written consent after they received the study description. Subjects denied any history of neurological or musculoskeletal injury. Demographic information is provided in Table \ref{Table1}.

\subsection{Data Analysis and EoP Calculation}
Using the OSP model of the dynamics of the hand \eqref{EoP_calc}, the EoP of the subject’s limb can be calculated in a given direction `$i$', muscle activation level `$m$', and interaction frequency `$\omega$':

\begin{equation} \label{EoP_calc}
    \xi_{i,m,\omega} = \frac{\int_{T_s}^{T_e} f_{i,m,\omega}(t)^T v_{i,m,\omega}(t)dt} {\int_{T_s}^{T_e} v_{i,m,\omega}(t)^T v_{i,m,\omega}(t)dt} 
\end{equation}

In \eqref{EoP_calc}, $\xi_{i,m,\omega}$ is the calculated EoP given the set of conditions `$i,m,\omega$', $f_{i,m,\omega}$ is the measured reactive force, and $v_{i,m,\omega}$ is the velocity of the hand, $T_s$ and $T_e$ start and end times of the last 5-second window from the corresponding 10-sec perturbation. The last five-second window is chosen to avoid the artifact caused by switching the perturbation direction. The mean \%MVC is also calculated along each direction for the same corresponding time range used to calculate the EoP for each test (LR, LS, HR, and HS). The resulting GMP maps are generated for each subject showing the effect of the interactive factors on the EoP. The box plot of all the EoPs for each test (LR, LS, HR, and HS) is plotted in Fig. \ref{eopbox} and discussed in Section IV. Additionally, the overall relationship between EoP and \%MVC is shown in Fig. \ref{scatter} and discussed later in Section IV. 

\section{Results}
Fig. \ref{maps} shows each subject’s corresponding Geometric Myo-Passivity Map for the four test settings (LR, LS, HR, and HS). The blue dots in each spider plot represent the identified EoP of the biomechanics with regard to the low-frequency perturbation. The red dots represent the identified EoP for the high-frequency perturbation. The top row shows the GMP maps during the relaxed condition, which reflects a low \%MVC. And the bottom row shows the GMP maps during the stiff condition, which reflects maintaining a 40\% MVC controlled through the visual myo-feedback. As can be seen, under low-frequency perturbation, the identified EoP values are significantly higher than the EoP values identified during high-frequency perturbation. This observation has been made in each direction of perturbation across all subjects and at the two muscle activation levels. Moreover, when the subject’s muscle activation is increased (from the relaxed to the stiff condition), the EoP in nearly all directions increases significantly, some even reaching twice their relaxed EoP values. The phenomenon reflected that the intrinsic biomechanical characteristic of the human limb is a multi-variable dependent factor that relies not only on the level of muscle activation but also on the frequency of the interaction and the directional geometry of interaction.

\begin{figure}[!h]
\centerline{\includegraphics[width=0.4\textwidth]{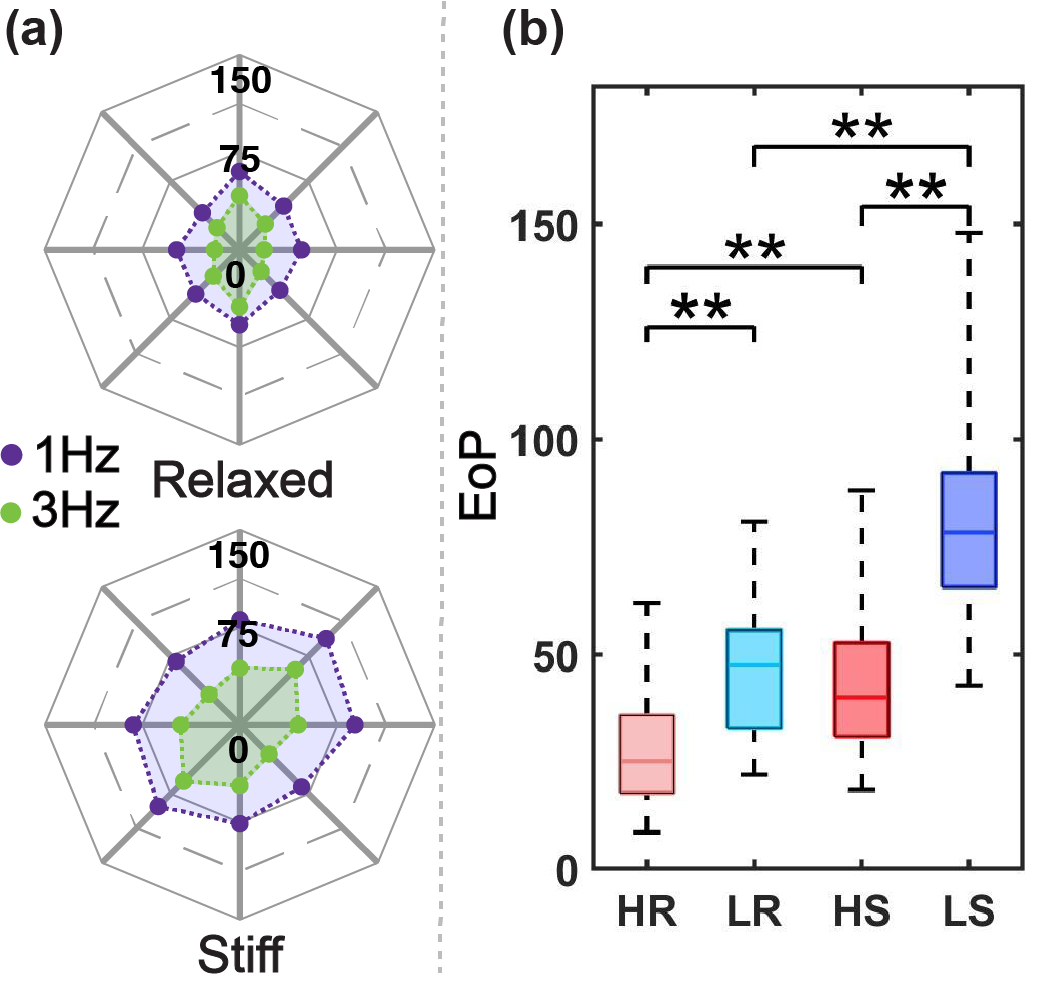}}
\caption{(a) Median GMP map of all five subjects during relaxed muscle co-activation and stiff muscle co-activation. (b) Box Plot distribution of the EoP values for the four tests (HR, LR, HS, and LS).}
\label{eopbox}
\end{figure}

Fig. \ref{eopbox}(a) shows the median GMP map across all five subjects during relaxed and stiff conditions. As shown, the median GMP map follows the same observations mentioned above (i.e., the low-frequency profile is larger than the high-frequency profile, and the GMP map of high muscle-activation is larger than the relaxed GMP map). In order to evaluate the statistical significance of the observed behavior of the median GMP map, a statistical analysis was also conducted, as can be seen in Fig. \ref{eopbox}(b). In Fig. \ref{eopbox}(b) and the rest of the paper, the tests are color-coded. For this, the low-intensity red is considered for HR, the low-intensity blue is considered for LR, the high-intensity red is considered for HS, and the high-intensity blue is used for LS. In order to calculate the significance, the Wilcoxon Sign Rank Test \cite{stat1} is used since the groups were found to be non-normal. The normality was tested using the Kolmogorov-Smirnov Normality Test \cite{stat2}. The significance threshold is considered to be 0.05 for the p-value `$p$'. The significance of $p<0.05$ is indicated by `*'; also, the significance $p<0.001$ is indicated by `**' in this paper.

Fig. \ref{eopbox}(b) shows the box plot distribution of the EoP values for all subjects during the four test settings (LR, LS, HR, and HS). In each box plot of Fig. \ref{eopbox}(b), there are forty EoP values which correspond to  five subjects and eight directions. In this paper, we specifically investigated the statistical difference between high and low-frequency groups and between high and low-muscle co-activation groups. As shown, the EoP during low-frequency perturbations is statistically larger than the EoP during high-frequency perturbations. Likewise, the  EoP during high muscle activation is statistically larger than the EoP during low muscle activation.

\begin{figure}[!h]
\vspace{0.2cm}\centerline{\includegraphics[width=0.48\textwidth]{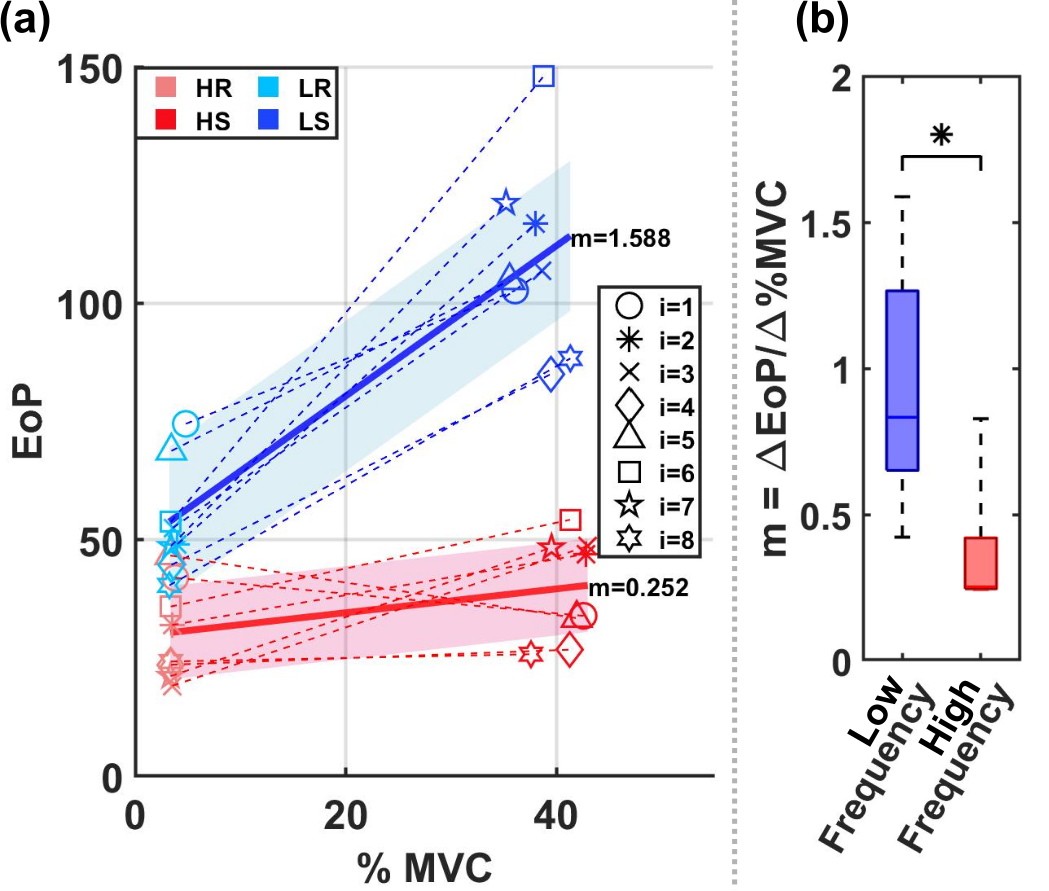}}
\caption{(a) Scatter plot showing the EoP vs \%MVC in each direction during the four tests of one subject. Eight markers are used to define eight directions of interaction. The trend line of the HR-HS group and the LR-LS group and the resulting slope are shown. (b) Box plot distribution of the low-frequency and high-frequency trend line slopes for all five subjects.}\vspace{-0.4cm}
\label{scatter}
\end{figure}

As mentioned previously, we hypothesize that the relationship between EoP and \%MVC is modulated by the changes in the frequency of interaction. As a result, Fig. \ref{scatter}(a) shows EoP versus \%MVC scatter plot for one subject (as an example) during the four tests. Also, Fig. \ref{scatter}(b) shows the box plot distribution of  $\Delta EoP/\Delta\%MVC$ for low-frequency versus high-frequency interaction of all five subjects. To clarify, to evaluate the relationship between the change of the EoP with respect to the change in the \%MVC,  a trend line can be calculated for each subject using the first-order least-squares method on all observations of EoP versus \%MVC. The trend line is the line of best fit representing the change. The trend lines can be seen for one subject in Fig. 5(a), in which the solid blue line shows the trend line for LS-LR tests, and solid red shows the trend line for HS-HR tests. The slopes of the trend lines (calculated as $\Delta EoP/\Delta\%MVC$) of all subjects form a distribution for low frequency and one for high frequency. 

As mentioned in Fig. \ref{scatter}(a), the high-frequency trend line (i.e., HR-HS) is shown in red, with the standard deviation shown as a shaded red region. Likewise, the low-frequency trend line (LR-LS)  is shown in blue, with the standard deviation shown as a shaded blue region. As can be seen, the slope of the low-frequency trend line is much larger than the slope of the high-frequency trend line. This observation indicates that during low-frequency interaction, a change in the \%MVC will greatly impact the change in the EoP. In other words, during low-frequency interaction, the level of muscle activation is one of the crucial factors in the resulting biomechanic energy absorption capability. This observation highlights the compounding factor of frequency and muscle co-activation. In order to statistically evaluate this phenomenon, a box plot of the trend line slopes is generated for all five subjects and shown in Fig. \ref{scatter}(b). Similar to Fig. \ref{eopbox}(b), the Wilcoxon Sign Rank Test is used to measure the significance between the low-frequency and high-frequency slopes due to the non-normality of the distributions. The results show that the slope `$\Delta EoP/\Delta\%MVC$' during low-frequency interaction is statistically larger than the slope during high-frequency interaction across all five subjects. Based on this observation, the frequency of the interaction during pHRI plays a modulating role which influences the effect muscle activation will have on the EoP. Particularly, during low-frequency interaction, muscle activation significantly influences the resulting EoP. However, the influence is not as strong for high-frequency interaction.

The results mentioned above highlight the possibility of estimating the EoP of human biomechanics based on three compounding factors (i.e., EMG co-activation, frequency of interaction, and geometrical direction of task). The estimation of EoP will allow stabilizers to calculate the margin of passivity in the closed-loop system. This estimation can be exploited to reduce the conservatism of control, minimizing the need for synthetic dissipation and maximizing the fidelity of force reflection, thus optimizing energy flow between human biomechanics and robot mechanics in pHRI. The results were highly consistent for all included participants of the study. However, the study was limited by the number of participants, which is the core line of future work.

\section{Conclusion}

In this paper, the frequency-based geometric myopassivity map was proposed that takes into account the interactional frequency besides operational geometric direction and level of muscle activity to estimate the energetic capacitance of human biomechanics in absorbing the interaction energy. The design relaxes the classic assumption of linearity on human biomechanics and can be used to formulate adaptive control schemes for pHRI to maximize the margin of passivity while maintaining a high performance (such as motion tracking or force reflection in haptics and telerobotics systems). In this study, five healthy subjects participated. The results show that the frequency of interaction is a modulating factor on the biomechanical EoP due to two important observations. First, the frequency of interaction impacts the resulting biomechanical energy absorption capability even when the muscle activity and perturbation direction are regulated. Second, the level of the interactive frequency modulates the contribution that the muscle activity can have on the resulting biomechanical energy absorption capability. Specifically, during low-frequency interaction, muscle activity plays a significantly higher role on the EoP when compared with high-frequency interactions. The results were statistically evaluated using the Wilcoxon sign rank test. The application domain for the proposed GMP map includes a wide range of pHRI systems, such as haptic systems, telerobotic systems,  rehabilitation robotics, and exoskeletons, specifically when there is a need for high-fidelity force feedback. In the future line of research, we will (a) increase the number of subjects recruited in the study and (b) design and implement a new family of stabilizers which can optimize the interaction based on real-time observations of the GMP map.

\bibliographystyle{IEEEtran}
\bibliography{ref}

\end{document}